\documentclass[twoside,11pt]{article}
\usepackage{jair, theapa, rawfonts}

\usepackage[utf8]{inputenc}
\usepackage{url}
\usepackage{amsmath}
\usepackage{amssymb}
\usepackage{graphicx}
\usepackage{adjustbox}
\usepackage{multirow}
\usepackage{pifont}

\usepackage{xcolor} 
\newtheorem{definition}{Definition}

\newtheorem{example}{Example}

\ShortHeadings{Computational Argumentation-based Chatbots: a Survey}{Castagna, K\"okciyan, Sassoon, Parsons, \& Sklar}
\firstpageno{1}

\title{Computational Argumentation-based Chatbots: a Survey}
\author{\name Federico Castagna \email federico.castagna@brunel.ac.uk \\
       \addr Brunel University London, Kingston Lane,\\ London, UB8 3PH, United Kingdom
       \AND
       \name Nadin K\"okciyan \email nadin.kokciyan@ed.ac.uk \\
       \addr University of Edinburgh, Crichton St,\\ Edinburgh EH8 9AB, United Kingdom
       \AND
       \name Isabel Sassoon \email isabel.sassoon@brunel.ac.uk \\
       \addr Brunel University London, Kingston Lane,\\ London, UB8 3PH, United Kingdom
       \AND
       \name Simon Parsons \email sparsons@lincoln.ac.uk\\
       \addr University of Lincoln, Brayford Pool,\\ Lincoln, LN6 7TS, United Kingdom
       \AND
       \name Elizabeth Sklar \email esklar@lincoln.ac.uk\\
       \addr University of Lincoln, Brayford Pool,\\ Lincoln, LN6 7TS, United Kingdom}

\begin{document}

\maketitle
\begin{abstract}
 Chatbots are conversational software applications designed to interact dialectically with users for a plethora of different purposes. Surprisingly, these colloquial agents have only recently been coupled with computational models of arguments~(i.e. computational argumentation), whose aim is to formalise, in a  machine-readable format, the ordinary exchange of information that characterises human communications. Chatbots may employ argumentation with different degrees and in a variety of manners. The present survey sifts through the literature to review papers concerning this kind of argumentation-based bot, drawing conclusions about the benefits and drawbacks that this approach entails in comparison with standard chatbots, while also envisaging possible future development and integration with the Transformer-based architecture and state-of-the-art Large Language models. 
\end{abstract}

\section{Introduction}
Chatbots are conversational software applications designed to mimic human discourse mostly to enable automated online guidance and support \shortcite{caldarini2022literature}. These computer programs generate responses based on given inputs, producing replies via text or speech format \shortcite{sojasingarayar2020seq2seq,bala2017chat}. In addition, to be defined as such, chatbots must satisfy specific functions. As colloquial agents, they need to be able to understand the user~(\textit{comprehension}), have access to a knowledge base (\textit{competence}) and provide an `anthropomorphic effect' to increase the users' trust (\textit{presence}) \shortcite{cahn2017chatbot,sansonnet2006architecture}. Nowadays, these bots represent familiar tools that exist in our lives in the form of virtual agents. Their assistance ranges from answering inquiries to e-commerce, from information retrieval to educational tasks, and from developing new industrial solutions \cite{dale_2016} to connecting smart objects \cite{Kar2016}. The manifold investments of the past decade, the technological advancements (from both software and hardware viewpoints), and the development of more efficient Machine Learning (ML) models, including the latest Transformer-based architecture \shortcite{vaswani2017attention}, have contributed to the steady growth of the research field of chatbot design and implementation. Many steps forward have been taken since the release of \emph{ELIZA} around sixty years ago, which is widely considered to be the first conversational agent~\cite{weizenbaum1966eliza}. 

The investigation of computational models of arguments in relation to chatbots has only recently received attention from researchers. Computational argumentation~\cite{ArgAI} has been applied in Artificial Intelligence~(AI) as a mechanism for reasoning in which conclusions are drawn from evidence that supports the conclusions.  Being an intuitive (i.e. closer to everyday human dialectical interplay), yet formal, approach for modelling  conflicting information occurring during exchange of arguments, computational argumentation should be qualified as a highly appropriate methodology to enhance current bot behaviours. The benefits from such a combination include: more natural discourse, response coherence and strategical conveyance of information. Evaluating argumentation semantics would also provide the rationale for positing replies in a more transparent way than the black-box Large Language models (LLMs) employed in today's state-of-the-art conversational agents.  
 In recent years, cutting-edge technologies have produced implementations, such as the various versions of ChatGPT\footnote{\url{https://chat.openai.com/}},  which currently outperform argumentation-based conversational agents. Nonetheless, taking a closer look---as we do here---shows that there is plenty of room for improvement for these recent advanced models, and integration with the computational argumentation formalism may solve their present shortcomings~(e.g. lack of explainability), thus potentially initiating a new generation of chatbots. To the best of our knowledge, this is the first survey that combines computational argumentation and chatbots\footnote{Notice that, for simplicity, we are often going to prefer the terminology `argumentation-based chatbot' rather than `computational argumentation-based chatbot', although the meaning will remain the same.}. Our main contribution involves an extensive examination of the relevant literature and the subsequent findings that can be drawn from such analysis.

The paper is structured as follows. We first start by introducing background information in Section~\ref{sec:bg} about the essential theoretical notions involved. In Section~\ref{sec:meth}, we then discuss the methodology adopted for reviewing the relevant articles. A thorough classification and analysis of conversational agents leveraging computational argumentation is given in Section~\ref{sec:chatbots}. Section~\ref{sec:discussion} illustrates a comprehensive examination of the paper's findings and potential future directions of the argumentation-based chatbot research field, and Section~\ref{sec:conc} concludes the survey with final remarks.

\section{Background}\label{sec:bg}
The following background covers a concise summary of computational argumentation, along with a short overview of the history, classification and main features of chatbots. The information provided will prove useful for the analysis undertaken in the next sections, where each conversational agent will be classified according to the specific argumentation employment presented herein.

\subsection{Computational Argumentation}
The term `computational model of arguments' encompasses a wide range of different approaches, each of which revolves around the notion of arguments and their employment. The resulting research field, whose roots can be traced back to Pollock's and Dung's systematical account of arguments~\cite{pollock1987defeasible,dung1995acceptability}, constitutes a rich interdisciplinary environment comprising subjects such as philosophy \shortcite{walton1990reasoning,mercier2011humans}, jurisprudence \shortcite{bench2009argumentation}, linguistics \shortcite{lawrence2020argument}, formal logic \shortcite{lin1989argument} and game theory \shortcite{rahwan2009argumentation}. Within the scope of computational argumentation, it is possible to identify two main research goals: (a) understand argumentation as a cognitive phenomenon via computer program modelling; and (b) support the development of human-computer interaction by means of argumentation-related activities \shortcite{prakken2020computational,sep-argument}. According to Dung's paradigm \cite{dung1995acceptability}, arguments are considered suitable means to formalise non-monotonic reasoning, especially when showing how humans handle conflicting information in a dialectical way. The core notion of such an approach is underpinned by the definition of an argumentation framework, where arguments are intended as abstract entities:
\begin{definition}[Abstract AFs~\cite{dung1995acceptability}]
An argumentation framework \emph{(AF)} is a pair: $\mbox{AF} = \langle \mbox{AR}, \mathcal{C} \rangle$ where AR is a set of arguments, and $\mathcal{C}$ is the `attack' binary relation on AR, i.e. $\mathcal{C}$ $\subseteq$ AR $\times$ AR.
\end{definition}
AFs can be rendered as graphs where each node is an argument, and every directed edge connects the conflicting arguments of the framework. 
The idea conveyed by this formalism is that correct reasoning is rendered via the acceptability of a statement: an argument is \emph{justified} only if it is defended against any counterarguments.
\begin{definition}[Semantics for Abstract AFs~\cite{dung1995acceptability}]
\label{Dung's semantics}
Let $\mbox{AF}=\langle \mbox{AR}, \mathcal{C} \rangle$, and let $\mathcal{S} \subseteq$ AR be a set of arguments. Let also $($X,Y$) \in \mathcal{C}$ denote the conflict existing between an argument $X$ and its target $Y$:
\begin{itemize}
\item $\mathcal{S}$ is \emph{conflict-free} iff $\forall X, Y \in \mathcal{S}$: $(X, Y) \notin \mathcal{C}$;
\item $X \in$ AR is acceptable w.r.t. $\mathcal{S}$ iff $\forall Y \in$ AR such that $(Y, X) \in \mathcal{C}$: $\exists Z \in \mathcal{S}$ such that $(Z, Y) \in \mathcal{C}$;
\item A conflict-free extension $\mathcal{S}$ is an \emph{admissible} extension iff $X \in \mathcal{S}$ implies $X$ is acceptable w.r.t. $\mathcal{S}$;
\item An admissible extension $\mathcal{S}$ is a \emph{complete} extension iff $\forall X \in$ AR: $X$ is acceptable w.r.t. $\mathcal{S}$ implies $X \in \mathcal{S}$. The minimal complete extension $($with respect to set inclusion$)$ is called the \emph{grounded extension}, whereas a maximal complete extension $($with respect to set inclusion$)$ is called a \emph{preferred extension};
\item A \emph{stable extension} $\mathcal{S}$ is such that iff $\forall Y \in$ AR, if $Y \notin \mathcal{S}$, then $\exists X \in \mathcal{S}$ such that $(X, Y) \in \mathcal{C}$. 
\end{itemize}
\end{definition}

Furthermore, AFs can be instantiated by the formulae of some logical language. These instantiations paved the way for a plethora of different studies (e.g.,~\cite{besnard2008elements,modgil2013general,toni2014tutorial}) concerning the so-called \emph{structured argumentation}, as opposed to the previously introduced abstract approach. The internal structure of an argument is usually composed of (one or more) premises, a conclusion and a set of inference rules (e.g. strict or defeasible) connecting premises to the conclusion. The same semantics described above can then be used to evaluate  structured argumentation frameworks and compute justified arguments. 
\begin{example} 
Let us consider the abstract AF depicted in Figure \ref{fig:AFexample}. Then, according to the semantics described in Definition \ref{Dung's semantics}, we can identify the following extensions:
\begin{itemize}
    \item[] \emph{admissible} = $\emptyset$, $\{a\}$, $\{b\}$, $\{e\}$, $\{a,e\}$, $\{b,e\};$
    \item[] \emph{complete} = $\{e\}$, $\{a,e\}$, $\{b,e\};$
    \item[] \emph{grounded} = $\{e\};$
    \item[] \emph{preferred} = $\{a,e\}$, $\{b,e\};$
    \item[] \emph{stable} = $\{a,e\}$, $\{b,e\}.$
\end{itemize}
    \begin{figure}[ht!]
    \centering
    \includegraphics[width=0.6\linewidth]{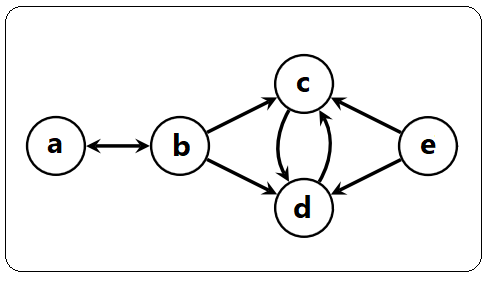}
    \caption{\footnotesize{An abstract argumentation framework.}}
    \label{fig:AFexample}
\end{figure}
\end{example}

\subsubsection{Argument Mining} \label{sec:AM}
Argument(ation) mining has been defined as \textit{``the general task of analyzing discourse on the pragmatics level and applying a certain argumentation theory to model and automatically analyze the data at hand''} \cite{habernal2017argumentation}.
\emph{Argument mining} (AM) can be considered the research area aimed at detecting natural language arguments and their relations in text, with the final goal of providing machine-processable structured data for computational models of argument \cite{cabrio2018five}. As depicted in Figure~\ref{fig:arg-emp-schematic}, an AM pipeline consists of two main stages: arguments' extraction and relations' prediction. We could delineate the AM framework by listing the tasks, in increasing order of complexity, that constitute such a framework. In short, moving from a preliminary textual segmentation and a classification of such elements as argumentative or not, it will then be possible to identify the single argument components (such as premises, claim, major claim,
evidence, etc. \shortcite{mayer2020transformer}). The following steps envisage the recognition of clausal properties and relational properties with respect to the previously detected argument components \cite{lawrence2020argument}. In particular, Saadat-Yazdi, Pan and K\"okciyan show how the use of external commonsense knowledge helps in identifying relations among arguments by uncovering implicit inferences~\shortcite{saadat2023uncovering}.
Some of the models proposed in the literature include Long-Short Term Memory~(LSTM) models~\cite{cocarascu_identifying_2017}, pre-trained transformers~\shortcite{ruiz-dolz_transformer-based_2021,saadat2023uncovering} and logical rule-based systems~ \shortcite{jo_classifying_2021}. Overall, AM is useful in enabling the generation of an argumentation framework, or graph, from the mined corpus of texts. We now provide a more concrete analysis of the arguments' extraction stage within the AM pipeline.
\begin{example}
Inspired by the political debate example illustrated in \shortcite{cabrio2018five}, we introduce an example to show how one can identify single arguments by following two distinct steps: (S1) the detection of argument components, such as premises and claims, and (S2) the recognition of their specific textual boundaries via the exclusion of any irrelevant words. In the following, we show how S1 and S2 could be applied to an example about the use of solar energy to extract an argument (\textit{Arg}). Note that \textbf{(C)} and \textbf{(P)} distinguish conclusion from premises, whereas the bold and underlined fonts identify their respective boundaries. 
\begin{itemize}
    \item[(S1)] ``She talks about solar panels. We invested in a solar company, our country. That was a disaster \textbf{(C)}. They lost plenty of money on that one \textbf{(P)}. Now, look, I'm a great believer in all forms of energy \textbf{(P)}, but we're putting a lot of people out of work \textbf{(P)}.''
    \item[(S2)] ``She talks about solar panels. We invested in a solar company, our country. \textbf{That was a disaster.} \underline{They lost plenty of money on that one}. Now, look, \underline{I'm a great believer} \underline{in all forms of energy}, but \underline{we're putting a lot of people out of work}.''
    \item[(Arg)] [Since] they lost plenty of money on that one, [even though] I’m a great believer in all forms of energy, we’re [nonetheless] putting a lot of people out of work. [We can then conclude] that was a disaster. 
\end{itemize} 
    \begin{figure}[h!]
    \centering
    \includegraphics[width=0.9\linewidth]{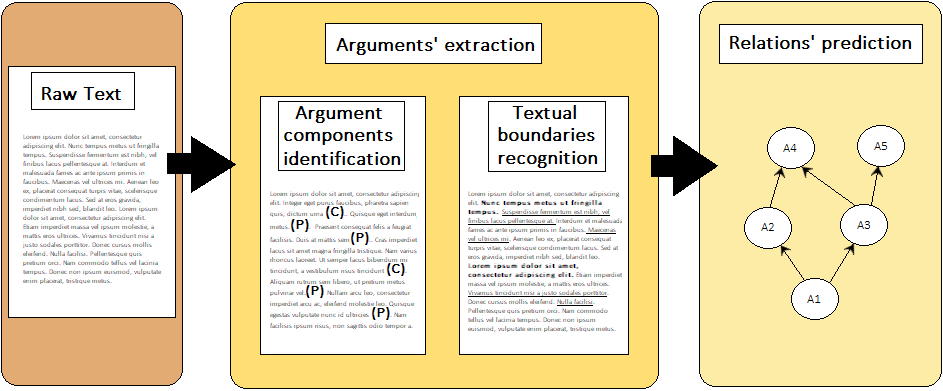}
    \caption{\footnotesize{Example of an argumentation mining pipeline.}}
    \label{fig:arg-pipeline}
\end{figure}
\end{example}

\subsubsection{Argument Schemes} \label{sec:AS}
Argument schemes (AS) have been extensively investigated and employed in the AI literature as a way to directly convey presumptive reasoning in multi-agent interactions (e.g. \shortcite{atkinson2006argumentation,tolchinsky2012deliberation,grando2013argumentation,kokciyan-sassoon-18,kokciyan2021applying}). Each AS is characterized by a unique set of critical questions (CQs), rendered as attacking arguments, whose purpose is to establish the validity of the scheme instantiations (which can then be evaluated by semantically computing their acceptability).
Although the literature presents diverse classification systems for argument schemes (e.g. \shortcite{walton2008argumentation,walton2015classification,wagemans2016constructing}), they all share the idea that such schemes constitute reasoning patterns that may be harnessed to structure natural language text into rational and coherent arguments, thus generating systematic elements of dialogue. 
\begin{example}
   As an example of AS in the healthcare domain, consider the argument scheme for proposed treatment (ASPT), as rendered in \shortcite{sassoon2021argumentation}, and the respective critical questions: the validity of any potential ASPT instantiation depends upon the answers given to each critical question.
\begin{table}[h]
\label{ASPT table}
\begin{small}
\[
\begin{tabular}{|l|}
\hline
\textbf{ASPT}\\
\hline
\\

\emph{Premise} $:$ \mbox{Given the patient's fact Ft}\\
\emph{Premise} $:$ \mbox{In order to realise goal G}\\
\emph{Premise} $:$ \mbox{Treatment T promotes goal G}\\
\par\noindent\rule{7cm}{0.4pt}\\
\emph{Conclusion} $:$ \mbox{Treatment T should be considered}\\
\hline
\end{tabular}
 \]

\begin{itemize}
\item[\textbf{CQ1}:]Has treatment T been unsuccessfully used on the patient in the past?
\item[\textbf{CQ2}:]Has treatment T caused side effects for the patient?
\item[\textbf{CQ3}:]Given the patient's fact Ft, are there counter-indications to treatment
T?
\item[\textbf{CQ4}:]Are there alternative Actions to achieve the same goal G?
\end{itemize}
\end{small}
\end{table} 
\end{example}

Finally, although the concept was developed for different purposes, the importance of argument schemes has found uptake within the computational argumentation community \shortcite{visser2018revisiting} also for textual mining tasks \cite{walton2012argument}.

\subsubsection{Argumentation Reasoning Engine}
\label{sec:Arg Reasoning Engine}
One of the main purposes of computational argumentation is to enable the resolution of conflicting knowledge, thus allowing for a selection of the most appropriate (i.e. justified) pieces of information.
\textit{``A decision is a choice between competing beliefs about the world or between alternative courses of action. [...] Inference processes generate arguments for and against each candidate [belief or action]. Decision making then ranks and evaluates candidates based on the underlying arguments and selects one candidate as the final
decision. Finally, the decision commits to a
new belief about a situation, or an intention
to act in a particular way.''} \shortcite{fox2007argumentation}.
Decision-making processes can be encoded as problems whose solutions are rendered by the computation and evaluation of AFs: an argumentation engine is essentially a reasoning tool driven by the same logic. The resulting acceptable entities provide a compelling rationale for and against a given choice, while also leaving space for further deliberations \shortcite{dix2009research}.  
Such an argumentative decision-making apparatus can be a useful addition to any real-world software application concerning defeasible reasoning, as advocated by the comprehensive study of Bryant and Krause \citeyear{bryant2008review}.
We can distinguish two kinds of reasoning engines based on computational argumentation: 
\begin{itemize}
    \item `Solvers', i.e. specialized pieces of software that encode and provide answers to distinct algorithmic problems. In particular, they address computational argumentation-related reasoning challenges according to a chosen semantics $\sigma$: for example, the enumeration of $\sigma$-extensions in the AF and the credulous and sceptical membership of a specific argument to at least one (credulous) or each (sceptical) $\sigma$-extensions (e.g. \emph{AFGCN} \shortcite{malmqvist2021afgcn}, \emph{A-Folio DPDB} \shortcite{fichte2021afolio}, \emph{ASPARTIX-V21} \shortcite{dvorak2021aspartixv21}, \emph{ConArg} \shortcite{bistarelli2021conarg}, \emph{FUDGE} \shortcite{thimm2021fudge}, \emph{HARPER++} \shortcite{thimm2021harper+},  \emph{MatrixX} \shortcite{maximilian2021matrix}, \emph{$\mu$-toksia} \shortcite{niskanen2021tosia}, \emph{PYGLAF} \shortcite{alviano2021pyglaf}).
    \item `Panoptic Engines', i.e. solvers designed to implement additional functionalities and customisation tools (e.g. \emph{ArguLab} \shortcite{podlaszewski2011implementation}, \emph{ArgTrust} \shortcite{tang2012argumentation}, \emph{Argue tuProlog} \shortcite{bryant2006argue}, \emph{IACAS} \shortcite{vreeswijk1994iacas}, \emph{CaSAPI} \shortcite{gartner2007casapi}, \emph{Prengine} \shortcite{hung2017inference}, \emph{PyArg} \shortcite{borg2022pyarg}, \emph{NEXAS} \shortcite{dachselt2022nexas}).
\end{itemize}

\begin{example}
    The \emph{ASP-Solver} ASPARTIX is an example of such an argumentation-driven reasoning engine. Starting from an AF as input, the Answer-Set-Programming solver will output the result of the specified reasoning task given a particular semantic (both encoded as ASP rules).
    \begin{figure}[h!]
    \centering
    \includegraphics[width=0.7\linewidth]{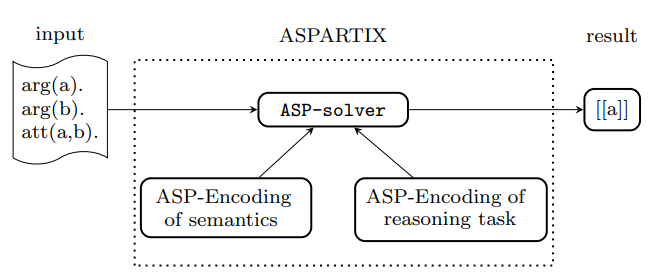}
    \caption{\footnotesize{Example of an argumentation reasoning engine architecture \shortcite{dvovrak2020aspartix}.}}
    \label{fig:aspartix-example}
\end{figure}
\end{example}
It is worth mentioning that most of these engines also embed a planning component, which derives from their underlying employment of the AF formalism. Indeed, computing acceptable arguments enables `argumentative paths' that lead to the achievement of the predetermined goal by deciding among (possibly) multiple options. Following edges that connect justified nodes in an AF will exclude any potential rebuttals, thus ensuring a successful strategy. That is to say, each reasoning step, enclosed and rendered as an argument, is performed whilst having in mind the overall plan required for reaching a consistent decision.

\subsubsection{Argumentation-based Dialogues}
The view of computation as distributed cognition and interaction contributed to the rise of the multi-agent systems paradigm, where agents are intended as software entities capable of flexible autonomous action in dynamic and unpredictable domains~\shortcite{luck2005agent}. As a means of communication between such intelligent agents, formal dialogues were chosen due to their potential expressivity despite still being subject to specific restrictions~\cite{mcburney2009dialogue}. Argumentation-based dialogues are \emph{rule-governed
interactions} among participants (i.e. agents with their own beliefs, goals, desires and a limited amount of information regarding the other players) that take turns in making utterances. As shown in Table 1, these dialogues are usually categorized according to elements such as information possessed by the participants at the commencement of the interaction, their individual goals, and the knowledge and goals they share with other agents~\cite{walton1995commitment}.

\begin{table}[ht!]
\centering
\label{dialogue types}
\begin{adjustbox}{width=\textwidth}
    \centering
\begin{tabular}{l l l}
\hline
        &\\ 
    \LARGE{\textbf{Dialogue type}} &  \LARGE{\textbf{Description}} & \LARGE{\textbf{Example}}\\
    & \\
    \hline
    \Large{\textbf{Information-seeking}} & \Large{X seeks the answer to some question(s) from Y.} & \Large{\shortcite{hulstijn2000dialogue}}\\
     \hline
     \Large{\textbf{Inquiry}} & \Large{X and Y collaborate to answer some question(s).} & \Large{\shortcite{black2007generative}}\\ 
     \hline 
      \Large{\textbf{Persuasion}}& \Large{X seeks to persuade Y to accept a proposition.} & \Large{\shortcite{prakken2006formal}}\\
     \hline
      \Large{\textbf{Negotiation}} & \Large{X and Y bargain over the division of some scarce resources.} & \Large{\shortcite{mcburney2003dialogue}}\\ 
     \hline
     \Large{\textbf{Deliberation}} & \Large{X and Y collaborate to decide what actions should be adopted.} & \Large{\shortcite{mcburney2007eightfold}}\\
     \hline
     \Large{\textbf{Eristic}} & \Large{X and Y quarrel verbally to vent perceived grievances.} & \large{/}\\ 
     \hline
     \Large{\textbf{Verification}}& \Large{X wants to verify the beliefs of Y.} & \Large{\shortcite{cogan2005new}}\\ 
     \hline
      \Large{\textbf{Query}} & \Large{X challenges Y since it is interested in Y's arguments.} & \Large{\shortcite{cogan2005new}}\\
     \hline
     \Large{\textbf{Command}} & \Large{X tells Y what to do.} & \Large{\shortcite{girle1996commands}}\\
     \hline
     \Large{\textbf{Education}} & \Large{X wants to teach Y something.} & \Large{\shortcite{sklar2004towards}}\\
     \hline
     \Large{\textbf{Chance discovery}} & \Large{Ideas arise out of exchanges between X and Y.} & \Large{\shortcite{mcburney2001chance}}\\
     \hline
\end{tabular}
\end{adjustbox}
\caption{\footnotesize{Description of existing dialogue types}}
\end{table}
The selection and transitions between different dialogues can instead be rendered via a \emph{Control Layer}~\cite{mcburney2002games,sklar2015argumentation}, defined in terms of \emph{atomic dialogue types} and \emph{control dialogues}. The latter are meta-structures that have as their topics other dialogues and contribute to the management of the protocols combinations and their transitions. 

In general, the main components of argumentation-based dialogues can be identified as: (i) \emph{syntax}, which handles the availability of and interaction between utterances; (ii) \emph{semantics}, which differs according to the specific focus and final deployment of the dialogue; and (iii) \emph{pragmatics}, which accounts for those aspects of the language that do not involve considerations about truth and falsity (e.g. the illocutionary force of the utterances)~\cite{mcburney2013talking}. 

\subsection{Chatbots}
A chatbot must be able to parse the user input and interpret what it means before providing an appropriate response or output (and thus starting a `chat'). The way in which the bot elaborates the replies to be delivered depends upon its \emph{response architecture model}. Following the studies conducted in \shortcite{adamopoulou2020chatbots,singh2020survey,klopfenstein2017rise,codecademy1}, we can classify such models as:
\begin{itemize}
    \item \textbf{Rule-based} chatbots 
    employ the simplest response architecture structure. The bots’ replies are entirely predefined and returned to the user according to a series of rules. The internal model of such rule-based software can be thought of as a decision tree that has a clear set of possible outputs defined for each step in the dialogue. Usually, this category of conversational agents handles those kinds of interactions where the user has a number of pre-compiled options to choose from. As an example of rule-based colloquial agents, we can consider ELIZA \cite{weizenbaum1966eliza}: deemed by scholars as the first implementation of a chatbot, it operates by harnessing linguistic rules in combination with recognized keywords from the users' inputs. Further development in the area resulted in PARRY \shortcite{colby1971artificial}, a chatbot that improved ELIZA via a conversational strategy embedded to simulate a person with paranoia. Jabberwacky \cite{jabber} is also an instance of a rule-based bot that interacts through contextual pattern matching. It steadily expands its database by collecting tokens from previous conversations that occurred with different users.
    \item \textbf{Retrieval-based} chatbots  
    represent a more complex response architecture structure. The bots’ replies are pulled from an existing corpus of stored sentences. Machine Learning and Natural Language Processing (NLP) models are used to interpret the user input (operation divided into \emph{intent classification} and \emph{entity recognition}) and determine the most fitting response to retrieve. As an example of retrieval-based colloquial agents, we can consider A.L.I.C.E. \cite{wallace2009anatomy} developed using the Artificial Intelligence Markup Language (AIML) \cite{wallace2003elements}. Such a language comprises a class of data objects and partially describes the behaviour of computer programs that process them via stimulus-response templates. Furthermore, also IBM's Watson Assistant \cite{watson} and Microsoft's Cortana \cite{cortana} represent other instances of the retrieval-based model. The first parses input to find statistically relevant replies in its database by means of parallel algorithms. The second instead leverages the natural language processing capabilities of Tellme's Network (owned by Microsoft from 2007) and the Satori knowledge repository to provide responses \cite{cortana2}.
    \item \textbf{Generative} chatbots 
    represent the most convoluted response architecture structure. These bots are capable of formulating their own original responses based on the user input rather than relying on existing text. The deployment of Deep Learning models allows returning the appropriate response by calculating the likelihood of the next element(s) in a word sequence. However, training such models requires time, and it is not always clear what is used to produce replies, which may be repetitive or nonsensical. In addition, generative bots are not generally capable of accessing data other than what is embedded in their model parameters. One common approach to mitigate these problems is to combine both retrieval and generative operations in the chatbot \shortcite{roller2020recipes}. As an example of such a hybrid type of virtual assistant, we can consider Apple's Siri \cite{siri} and Amazon's Alexa \shortcite{alexa,lopatovska2019talk}. Both provide replies to users' questions (along with an additional wide array of possible functions) via Deep Learning procedures or delegating requests to a set of external providers (e.g. WolframAlpha \cite{alexa2}).
    \\\\\\\emph{Generative-LLMs.} Generative chatbots that hinge upon Large Language models (LLMs) deserve special mention, given recent interest in such models. The design and deployment of the Transformer architecture \shortcite{vaswani2017attention} determined a paradigm shift towards `pre-training' and `fine-tuning' learnings \shortcite{zhao2023survey}: scaling up pre-trained models led to the discovery of LLMs and their impressive capabilities \shortcite{brown2020language,touvron2023llama,anil2023palm}. Leveraging these new technologies, conversational agents such as the famous ChatGPT\footnote{Other remarkable examples are DialoGPT \shortcite{zhang2019dialogpt}, BlenderBot 3x \shortcite{xu2023improving}, Bard (\url{https://bard.google.com/}), Claude (\url{https://claude.ai/}), Llama 2-Chat \shortcite{touvron2023llama2}, Mistral-7b-Instruct \shortcite{jiang2023mistral}, and Zephyr-7b \shortcite{tunstall2023zephyr}.} prove to outperform most of the previous benchmarks and predecessors in information extraction tasks \shortcite{li2023evaluating}, natural language inference, question answering, dialogue tasks \shortcite{qin2023chatgpt} and machine translation \shortcite{jiao2023chatgpt}. That being said, LLMs and the chatbots based on them also suffer from a number of downsides including: faulty reasoning, inexplicable appearance of previously unknown abilities (phenomenon denoted as \emph{emergent abilities}\footnote{Emergent abilities constitute a controversial topic and some studies even argue against their existence \shortcite{schaeffer2023emergent}.}), nonsensical or unfaithful replies (i.e. \emph{hallucination}), biased and toxic communications, expensive training costs and high carbon footprint\footnote{Although it has been argued that the adoption of best practices in model training should reduce carbon dioxide emissions by 2030 \shortcite{patterson2022carbon}.}. Finally, it has also been shown how underlying models such as GPT-3 \shortcite{brown2020language} fall short of producing adequate and compelling arguments \shortcite{hinton2022persuasive}. However the outputs of such models may prove particularly suited to support argument mining operations, given carefully conditioned (or an increased number of) inputs \shortcite{de2023wish,chen2023exploring}.
\end{itemize}
\begin{figure}[h!]
    \centering
    \includegraphics[width=1\linewidth]{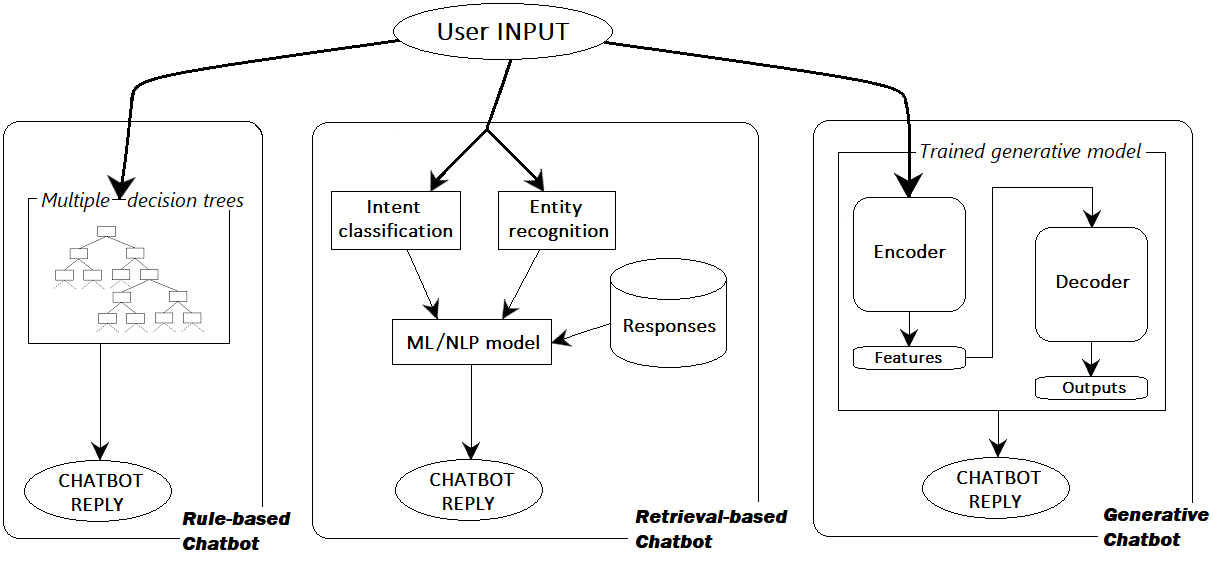}
    \caption{\footnotesize{Comparisons of different response architecture models}}
    \label{fig:comparisons}
\end{figure}
The different response architecture models and the corresponding high-level operations that characterise them are depicted in Figure \ref{fig:comparisons}. 
Notice, as previously anticipated, that is quite common for conversational agents to use a combination of different response models in order to produce optimal results. Furthermore, chatbots can be classified based on the conversation topics they are able to cover. \emph{Closed domain} ones (e.g. bots focused on customer assistance or e-commerce) are restricted to providing responses within a particular matter. Due to their specific area of competence, usually, these agents are very efficient in delivering good-quality discourses. On the other hand, \emph{open domain} chatbots (e.g. the previously referenced Apple's Siri, Amazon's Alexa, Meta's Llama 2-Chat, Google's Bard and OpenAI's ChatGPT, as well as Meena \shortcite{adiwardana2020towards}, Mitsuku \shortcite{worswick2018} and Microsoft's XiaoIce \shortcite{zhou2020xiaoice}) should be able to explore any range of conversation topics, similar to how a real-world human-to-human interaction would be. However, it is not straightforward to implement such bots, and they prove to be more prone to errors, incoherent responses\footnote{Notice these errors can have extreme and harmful consequences, such as a medical chatbot suggesting a patient to kill themselves. \cite{dangersOfAI}} or other issues similar to the aforementioned generative-LLMs.

\subsubsection{The Knowledge Base Acquisition}
Chatbots cannot automatically generate responses unless they are provided with a specific knowledge base from which those replies can be retrieved. This limitation involves every type of conversational agent and not only the retrieval-based, as one may think. Indeed, rule-based architecture requires hard coding of data into the scripts of the chatbot, whereas generative models necessitate a corpus of information to be trained upon. However, plenty of data collection is needed to obtain such knowledge bases and, usually, these datasets allow the chatbots to interact only on a restricted range of topics. In particular, anticipating a topic covered in the next sections, some argumentation-based chatbots are characterized by a knowledge base consisting of a set of arguments (alternatively, an argument graph) to collect which current approaches include argument mining from documents (e.g. \shortcite{cocarascu2019extracting,trautmann2020fine}) or hand coding of texts by researchers (e.g. \shortcite{cerutti2016pilot,rosenfeld2016strategical}).  
Nevertheless, these operations can be complicated tasks to achieve, especially if we need to handle only real-world arguments rather than artificial (i.e. computed) ones. That is to say, it may be difficult to retrieve high-quality arguments concerning a specific topic on the web, or it may also be problematic to distinguish between the person (and, thus, account for her attributes) who posited a specific claim. Questionnaires or personal interviews may provide a solution, although such solutions are expensive and require a large amount of human effort. Interestingly, studies such as \shortcite{chalaguine2018chatbot,chalaguine2018argument} proposed an alternative method to face this potential issue. The results of their research show how a chatbot, with little to no domain expertise, may elicit arguments and counterarguments from different users, thus automating the process of argument acquisition. This procedure, called \emph{argument harvesting} by the authors, allows for the generation of AFs that incorporate the knowledge base information over the required domain. Another alternative approach is provided by the work conducted in \cite{chalaguine2019knowledge} that describes how to acquire a large number of (high-quality) arguments in a graph structure using crowd-sourcing.

\section{Methodology}\label{sec:meth}
\begin{figure}[ht!]
    \centering
    \includegraphics[width=0.75\linewidth]{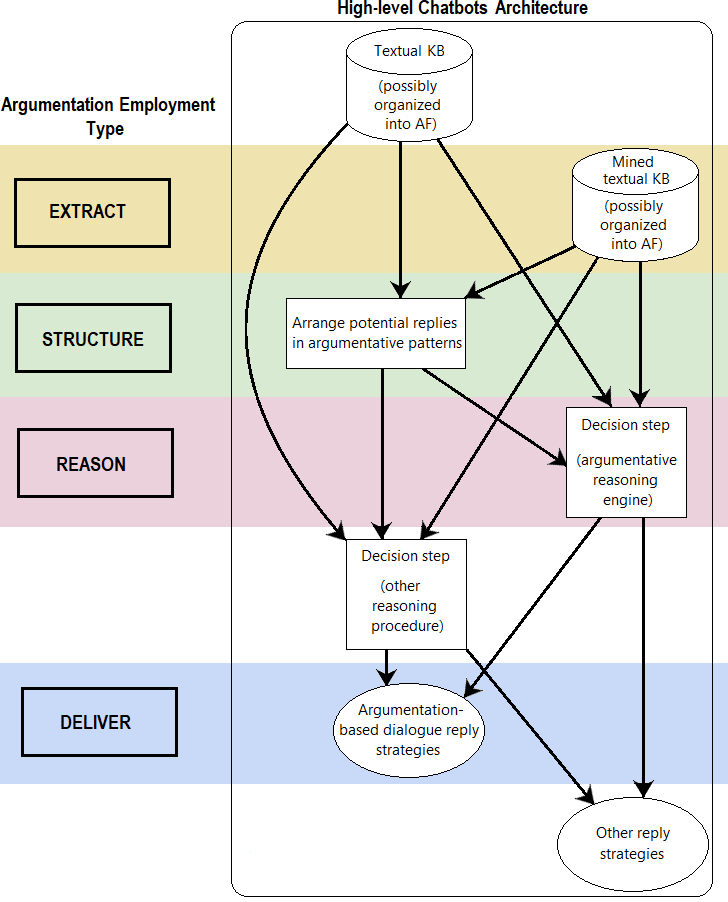}
    \caption{\footnotesize{Schematic argumentation employment within chatbots architecture.}}
    \label{fig:arg-emp-schematic}
\end{figure}

This survey hinges upon the collection and review of papers concerning argumentation-based chatbots. Before delving into the examination of our findings, it may be helpful to provide an uncontroversial definition of the subject of our investigation:
\begin{definition}[Computational Argumentation-based chatbots] We consider \emph{computational argumentation-based chatbots} those conversational agents that employ argumentative models to: (i) \emph{extract} textual data via argument mining tools, (ii) \emph{structure} information by means of argumentative templates, (iii) \emph{reason} with argument semantics and/or (iv) \emph{deliver} replies to users through argumentation-based dialogues. 
\end{definition}
A schematic representation of the argumentation employment types within a conversational agent architecture is provided in Figure~\ref{fig:arg-emp-schematic}. Here it is specified the level at which each aspect operates in the overall chatbot design. Argument mining enables the construction of a database for model training or a knowledge base (KB) by \emph{extracting} information from texts. KB data can be \emph{structured} into argumentative patterns, which may then be \emph{delivered} as argumentation-based dialogue replies to the interacting end-user after being selected through a \emph{reasoning} step (that usually involves argument semantics computation). 
Notice that we strictly selected only papers involving such aspects, avoiding any other articles pertaining to chatbots or meanings of argumentation that differ from those introduced in Section 2. For example, we did not consider the work of \cite{toniuc2017climebot} among the surveyed papers since it does not account for computational argumentation as described herein (albeit the presented \textit{textual entailments} relationship may be transformed into a sort of premises-conclusion argument dependency). A similar issue can be observed in \shortcite{altay2022scaling,kulatska2019arguebot} where, although there is a reference to a general notion of arguments and counterarguments, it does not correspond to the one provided in Section 2. On the other hand, we also excluded research such as \cite{chalaguine2018argument,chalaguine2018chatbot} since their focus is more on the automated collection of a corpus of arguments and counterarguments rather than the implementation of a conversational agent that delivers argumentation-based dialogues. Furthermore, the interactive recommender system of \shortcite{rago2018argumentation,rago2020argumentation}, which clarifies its recommendations through explanations, does not qualify as an argumentation-based chatbot either. That is because, although it makes use of a Bipolar/Tripolar AF\footnote{Bipolar Argumentation Frameworks, or Bipolar AFs, have been extensively introduced in \cite{cayrol2005acceptability}. Tripolar Argumentation Frameworks may instead be seen as (using the same words that appear in \shortcite{rago2018argumentation}): \textit{``instances of \emph{`tripolar frameworks'} as defined in \cite{gabbay2016logical} and of \emph{`generalised argumentation frameworks'} as defined in \shortcite{baroni2017abstract}.''}} to embed its underlying knowledge base, no reasoning, extraction, structure or delivery via computational models of argument (as interpreted in this paper) occurs. 

To clarify, in our survey, we did not restrict the search according to particular chatbot types, or their final scope, nor did we distinguish between different bot denominations (e.g. `argumentative dialogical agent', `dialogue manager', `automated persuasion system') or maturity of implementation, (e.g. fully-fledged or just sketched). Also, we did not account for specific time ranges and gathered articles independently of the year of publication. We have then analysed and organized the results in one concise comparative table  
(Table \ref{argumentation-based chatbots specifics})  
that displays the classifications and main features of each conversational agent. In particular, we listed all the reviewed chatbots and distinguished between each bot's final purpose (e.g. persuade, explain, inform), response architecture model (the prevalent one is recorded in case of multiple models), and conversation domain (for which we mostly considered the topic specified in the corresponding paper examples). Additional data comprise also the way in which computational argumentation has been employed within the chatbot architecture (i.e. extraction, structure, reason, deliver). Finally, we inspected the arranged information and discussed our main findings.
\vspace{1cm}
\begin{table}[ht!]
\centering
\begin{adjustbox}{width=\textwidth}
    \centering
    \begin{tabular}{lcccc}
         \hline
         \textbf{Paper(s)} & \textbf{Final} & \textbf{Response} & \textbf{Conversation} & \textbf{Argumentation}\\
          & \textbf{Purpose} & \textbf{Architecture} & \textbf{Domain} & \textbf{Employment}\\
         \hline 
         \cite{cocarascu2019extracting}&Explain&Rule-based&Movies reviews&Extract, Reason\\
         \hline
         (Slonim et al., 2021)&Debate&Retrieval-based&Unspecified&Extract\\
         &&&(semi-open domain)&\\
         \hline
         \cite{galitskyaffective}&&&&\\
         \cite{galitskysupporting}&Unspecified&Retrieval-based&Unspecified&Extract\\
         \cite{galitskyvalidation}&&&(closed domain)&\\
         \hline
         \cite{dignum2017creating}&Converse&Retrieval-based&Healthcare&Reason, Deliver\\
         \hline
         (Bistarelli et al., 2021)&Converse&Retrieval-based&Unspecified&Reason\\
         &&&(closed domain)&\\
         \hline
         (Castagna et al., 2022, 2023) &Explain&Retrieval-based&Healthcare&Reason, Structure\\
         \hline
         (Sassoon et al., 2019)&Explain&Retrieval-based&Healthcare&Reason, Structure,\\
         &&&&Deliver\\
         \hline
         (Fazzinga et al., 2021)&Inform&Retrieval-based&COVID-19 vaccine&Reason, Deliver\\
         \hline
         (Bex et al., 2016)&Inform&Retrieval-based&Fraud report&Reason, Deliver\\
         \hline
         \cite{sklar2015argumentation}&Explain&Retrieval-based&Treasure Hunt&Deliver\\
         \cite{sklar2018explanation}&&&Game&\\
         \hline
         \cite{rosenfeld2016strategical}&Persuade&Retrieval-based&Benefits of holding&Reason, Deliver\\
         &&&a Master's Degree&\\
         \hline
         (Chalaguine et al., 2019)&Persuade&Retrieval-based&Meat consumption&Deliver\\
         \hline
         \cite{chalaguine2020persuasive}&Persuade&Retrieval-based&UK university fees&Deliver\\
         \hline
         \cite{chalaguine2021addressing}&Persuade&Retrieval-based&COVID-19 vaccine&Deliver\\
         \hline
         \cite{hadoux2019comfort}&Persuade&Retrieval-based&Cycling in the city&Deliver\\
         \hline
         (Hadoux et al., 2021)&Persuade&Retrieval-based&UK university fees&Deliver\\
         \hline
         (Andrews et al., 2008)&Persuade&Retrieval-based&Desert survival&Deliver\\
         \hline
         \shortcite{guo2022using}
         &Persuade&Retrieval-based&Nuclear energy&Deliver\\
         \hline
         \shortcite{wambsganss2021arguebot}&Explain&Unspecified&Unspecified&Extract\\
         &&&(closed domain)&\\
         \hline
    \end{tabular}
\end{adjustbox}
\caption{Argumentation-based chatbots specifics}
\label{argumentation-based chatbots specifics}
\end{table}

\section{Argumentation-based Chatbots} \label{sec:chatbots}
This section covers a concise description of all the reviewed chatbots according to their specific argumentation employment. We first outline each argumentation-based category before providing an account of the conversational agents pertinent to the class. Note that it may be the case for a bot to present components that fulfil specific tasks (e.g. extract, structure, reasoning, and deliver) without exploiting computational argumentation. The fact that we are not detailing such components does not undermine their presence or effectiveness but reflects the choice of strictly conferring an argumentative scope to the survey. We conclude by highlighting the evaluations of such chatbots (if any) as presented in their respective papers.

\subsection{Argumentation-based Extraction}
Starting from a corpus of natural language texts, argument mining procedures allow for the extraction of arguments, and the classification of their relations, within such documents. The mined data can then be further processed and organized in AFs\footnote{We stipulate that constructing an AF from the utterances of an argumentation-based dialogue does not qualify as an `argumentation-based extraction'. That is because the arguments and the attacks (respectively supports) are already given and do not require further parsing.}, 
or simply be employed as replies according to the user's input. Unlike the latter, the former choice may lead to a reasoning operation upon the framework that will elicit specific output depending on the evaluation criteria of the captured semantics.  
For example, ADA, the argumentative dialogical agent introduced in~\cite{cocarascu2019extracting}, extracts arguments from movie review snippets and mines the relations subsisting among them. The acquired data is then utilized to construct a Quantitative Bipolar argumentation framework, QBAF (experimentally evaluated against three gradual semantics for QBAF: QuAD \shortcite{baroni2015automatic}, DF-QUAD \shortcite{rago2016discontinuity} and the Restricted Euler-based semantics (REB) \cite{amgoud2018evaluation}) upon which the conversational agent will instantiate the reply templates stored within its system. Those replies will thus be delivered to the interacting user when prompted for explanations about the selected movie recommendation.  

Another example of an argumentation-based extraction chatbot is rendered by the conversational agent developed in \shortcite{slonim2021autonomous} whose purpose is to challenge humans with competitive debates. After having preprocessed a corpus of 400 million newspaper articles in order to create an index of meaningful concepts, the bot mines for arguments thus obtaining claims and evidence related to the selected dispute. In this process, the agent identifies the relations occurring between the mined arguments and takes advantage of these data to prepare counterarguments against different stances on the debate topic. The replies posited by the bot will then be retrieved among the mined arguments, or the ones stored in a more general knowledge base, via a neural model. Notably, the interaction with the user occurs on a speech base and the speech-to-text conversion is performed by IBM's Watson\footnote{Once again, recall that we are emphasising the elements leveraging computational argumentation. Project Debater \shortcite{slonim2021autonomous} is a fully-fledged debating system, nonetheless, its employment of AM procedures is the only argumentation-related component, and this is why it is the one described.}.

A borderline case is constituted by the ArgueBot conversational agent \shortcite{wambsganss2021arguebot}. Developed as a learning tool for providing adaptive feedback on students' logic argumentation, ArgueBot (a bot deployed within the Slack platform\footnote{\url{https://slack.com/}}) hinges on a BERT \shortcite{devlin2018bert} classifier to perform AM operations on the user's textual input before providing tailored comments on their argumentative writing. Although the chatbot may be equipped with specific reply templates, its exact response architecture is unclear and remains unspecified by the authors.

Finally, on a more abstract level, the research discussed in \cite{galitskyvalidation,galitskyaffective,galitskysupporting} describes the deployment of specific argument mining approaches to chatbots. Here, the conversational agent constructs a communicative discourse tree from a subset of text by matching each fragment of the subset that has a verb to a verb signature. The subsequent application of classification models allows the bot to detect arguments and their relations and then leverage that information to provide replies according to the user input. In a nutshell, by resorting to in-depth rhetorical analysis, the chatbot accounts for multiple features of the argument (e.g. embedded affective aspects, consistency with the domain clauses, etc.), which results in more precise user-bot replies matches.

\subsection{Argumentation-based Reply Structures}
Chatbot replies can be structured according to the traditional argumentative format: a claim derived from a set of premises by means of particular inference rules. This approach includes argument schemes and general frameworks for structured argumentation (e.g. ASPIC$^{+}$, ABA, etc.). In general, the organization of data within such an argumentative pattern occurs before the generation of an AF and the computation of its semantics. However, it may also be convenient to arrange the bot responses using specific templates, regardless of a further semantic evaluation. Indeed, providing replies with a precise structure serves to highlight the rationale underpinning the argument claim and enhance the overall clarity of the discourse.   
As an example, we can consider the conversational agent presented in \shortcite{castagna2022providing,10.3389/frai.2023.1045614}, which may be seen as the final implementation of previous versions described in \shortcite{essers2018consult,kokciyan2019collaborative,balatsoukas2019user,chapman2019computational,balatsoukas2020wild,sassoon2020implementing,kokciyan2021applying,drake2022relationship}. Harnessing the novel Explanation-Question-Response, or EQR, argument scheme (first envisaged as a dialogue protocol and sketched in \cite{mcburney2021argument}), this bot delivers tailored justified recommendations within the healthcare domain, helping users self-manage their conditions. These recommendations embed
an additional layer of information: the rationale behind the instantiated scheme acceptability (i.e. its evaluation, automated via the ASPARTIX \shortcite{egly2008aspartix} engine, according to the considered argumentation framework). Additional replies provided by the chatbot will be structured by harnessing the argument scheme (and respective CQ) templates instantiated by the bot knowledge base. 

\subsection{Argumentation-based Reasoning}
As previously discussed, an argumentation engine can be employed as the underlying tool that drives a chatbot's \emph{reasoning} operations. In such a circumstance, regardless of the chosen framework (e.g. Abstract AFs \cite{dung1995acceptability}, Bipolar AFs \cite{cayrol2005acceptability}, Weighted Bipolar AFs \cite{rosenfeld2016strategical}, Quantitative Bipolar AFs \cite{cocarascu2019extracting}, Metalevel AFs \shortcite{kokciyan2021applying}, etc.), most of the decision-making processes involve the computation and semantic evaluation of the AF. Intuitively, starting from a knowledge base embedded in a set of arguments, the bot executes a reasoning procedure that usually results in a selection of acceptable arguments (which changes depending on the chosen semantics). When interacting with the user, the conversational agent will retrieve its replies, based upon the received input from its interlocutor, from the computed acceptable arguments. As such, we can generally assume that argumentation-based reasoning engines are intertwined with retrieval-based response architectures or hybrid models that include retrieval-based operations. For example, ArguBot \shortcite{bistarelli2021chatbot}, developed using Google DialogFlow\footnote{\url{https://cloud.google.com/dialogflow/docs/}}, employs ASPARTIX \shortcite{egly2008aspartix} to compute arguments from an underlying Bipolar AF, to support (\emph{pro-bot}) or challenge (\emph{con-bot}) the user's opinion about the topic of dialogue.

The conversational agent presented in \shortcite{fazzinga2021argumentative}\footnote{Subsequently embedded into a privacy-preserving dialogue system \shortcite{fazzinga2022privacy}.} retrieves its arguments from an underlying Bipolar AF as well, although it follows the semantics illustrated in \shortcite{ijcai2018p249}. The selected reply is then an argument acceptable with respect to an admissible extension computed over the overall framework, thus providing a strategy that also accounts for future developments of the chat. In addition, the bot is capable of formulating on-demand explanations about a particular reply, i.e. a sequence of natural language sentences that describes the facts supporting it, along with motivations against other possible conflicting arguments that the system discarded. 

In contrast, the chatbot outlined in~\cite{dignum2017creating} deploys computational argumentation as a means of evaluating completed phases of the ongoing dialogue, rather than starting with a previously generated AF. More precisely, an argument graph is constructed by incorporating the facts that emerge during the dialectical interaction with the user. Then, a formal assessment occurs by checking if those facts are members of acceptable extensions of the graph. Interestingly, this conversational agent harnesses social practices theory \shortcite{reckwitz2002toward,shove2012dynamics} to contextualise the conversation and provide useful background information that facilitates the user's input interpretation. A similar deployment of computational argumentation is envisaged in \shortcite{bex2016ai}, where an AI system that enhances the online report of trade frauds is outlined. A chatbot (the `dialogue manager') exchanges arguments with the user parties (both fraud victims and police) eliciting, if needed, more information about the ongoing case whilst building a knowledge graph. The acquired data will then enable the matching of the graph with a typical criminal scenario known by the police. Subsequently, formal argumentation semantics will drive the reasoning with scenarios and pieces of evidence (i.e. the `hybrid theory' \shortcite{bex2010hybrid,bex2015integrated}). 

Finally, the conversational agent (SPA) envisaged in \cite{rosenfeld2016strategical} also employs an argumentation-based reasoning engine. In particular, it embeds its knowledge base into a Weighted Bipolar AF (WBAF) and computes the argument that maximizes the framework evaluation function according to the user input. The score returned by the valuation function represents the reasoner’s ability to support that argument and defend it against potential attacks. The dialectical interaction with the user follows a strategical persuasion dialogue protocol (optimized via Monte Carlo Planning \cite{silver2010monte}) that might involve updating the argumentation frameworks of both the persuader and the persuadee. 

\subsection{Argumentation-based Reply Delivery}
Chatbots may handle and deliver their responses to the user interacting with them by leveraging the protocols of argumentation-based dialogues. Harnessing the dialogue logic, the conversational agent can optimize its strategy and utter only the arguments that prove to be necessary for achieving its final goal. In a way, we could identify the delivery phase as a `secondary reasoning step' where the bot chooses which arguments to move (strictly following the involved dialogue protocol instructions) among the ones available (possibly previously computed by the `primary engine' described by the reasoning phase). Notice that the arguments licensed in a dialogue protocol follow a more flexible definition than the standard ones provided in the abstract or structure argumentation approach: \emph{`` $[\ldots]$ it is the idea of dialogue as an exchange between two or more individuals, an exchange which captures features of what would be informally called an ``argument". That is, dialogue as the exchange of reasons $[$i.e. arguments$]$ for or against some matter''}\shortcite{black2021argumentation}.  

As an example, we could examine the work introduced in \shortcite{hadoux2021strategic}, which expands upon \shortcite{hadoux2019comfort,hunter2018towards,hunter2019towards} and depicts an overall framework for modelling beliefs and concerns 
in a persuasion dialogue. An implementation of such a framework is then envisaged via an automated persuasion system (APS), a software application aiming at convincing the interacting agent to accept some arguments. Following the asymmetric persuasion dialogue protocol illustrated therein (i.e. unlike the system, the user is restricted in choosing replies among the provided options), the proposed chatbot proves to be capable of identifying, within its knowledge base embedded in an argument graph, the most appropriate argument to posit. Essentially, the APS performs a Monte Carlo Tree Search coupled with a reward function to maximize the addressing of concerns (paired with the arguments of the graph) and the user's beliefs. 

Similarly, the bot presented in~\cite{chalaguine2020persuasive} aims at persuading the interlocutor via a free-text interaction where the user's inputs are matched (by vector rendering and cosine similarity) with the (crowdsourced) arguments of the graph representing the knowledge base. The chatbot trains a classifier to detect the most common concerns of the persuadee and employs it to select counterarguments that will produce a result more compelling than a random choice. If no argument similarity is detected, the conversational agent will resort to a default reply based on the user's concerns. Furthermore, the same authors presented an analogous architecture for a persuasion bot in~\cite{chalaguine2021addressing}, with the addition of a particular concern-argument graph. By incorporating the knowledge base within such a small graph, it can be proved that no large amount of data is needed to generate effective persuasive dialogues. Interestingly, a preliminary analysis of the impact (appeal) of arguments addressing the users' concerns in a persuasion dialogue performed by a chatbot has also been conducted by the same authors in~\shortcite{chalaguine2019impact}.
Another example of such a concern-based approach may be represented by Argumate, a chatbot designed to facilitate students’ production of persuasive statements \shortcite{guo2022using}. To provide appropriate suggestions, the bot retrieves its replies from an underlying argument graph, whose edges denote attack and support relations, via a concern identification method. Notice that the interactions between Argumate and the users occur both by typing and selecting predefined options.
\\\indent A common trait amongst all of the above argumentation-based conversational agents is that, although the corpus from which they extract their replies is organized as an argument graph, there is no interest in any particular acceptable semantics. That is to say, the knowledge base is organized and considered as a plain AF, where arguments and attacks are the only relevant features. In addition, most of these studies also account for a baseline chatbot which exploits a random strategy for selecting counterarguments from the available choices within the underlying knowledge base. The reason for this is to provide a means for comparing the developed bots which employ more fine-grained strategies for choosing their replies. 

Finally, one last conversational agent that focuses on the delivery of persuasion dialogues is the chatbot designed in~\shortcite{andrews2008argumentative}. Implemented harnessing the AIML markup language~\cite{wallace2003elements}, the bot comprises a planning component that searches over an argumentation model for the optimal dialectical path to pursue in order to persuade the user. The agent records the user's beliefs and updates this information whenever its interlocutor agrees/disagrees during the interaction. Such belief revision plays an important role in the strategic view of the chatbot. Moving towards different topics, the conversational agent implemented in~\shortcite{sassoon2019explainable}, within the context of explanation for wellness consultation, exploits multiple dialogue protocols (i.e. persuasion, deliberation and information seeking) whilst exchanging instantiations of acceptable argument schemes with its interlocutor. The adoption of diversified dialogue protocols (i.e. persuasion, inquiry and information seeking) characterises also the chatbot-equipped robot proposed in~\cite{sklar2015argumentation} and demonstrated in~\cite{azhar-sklar-ijrr:2017}. Retrieving the most appropriate argument constructed from its beliefs, an operation facilitated by the restricted options available to the user, the robot communicates with its human interlocutor in order to strategize about a treasure-hunting game. 

\subsection{Evaluation of the Chatbots}
Thus far, we have described the reviewed argumentation-based chatbots, primarily focusing on their features in relation to argumentation employment. However, some of those conversational agents have also been evaluated via specifically designed user studies\footnote{A different (and outdated) way of evaluating the capability of a conversational agent would be through a discussion with a human end-user: the more natural and seamless the interaction, the more effective the chatbot. 
The Turing Test (or \emph{Imitation Game}) is a proposal advanced by Alan Turing 
~\cite{turing1950computing} whose idea was to present some sort of test of a machine’s ability to exhibit intelligent behaviour equivalent to, or indistinguishable from, that of a human.  
Hinging on the Imitation Game, the Loebner Prize is a contest started in 1980 to award computer programs that are the most human-like, i.e. that perform the best in the Turing Test.  
The winner of the contest is the one that tricks a judge the highest percentage of the time, and Mitsuku is the chatbot that won the largest number of such prizes~\cite{worswick2013}.
The Loebner competition (considered defunct since 2020) has been subjected to a long list of criticisms. Among these, there was the alleged idea that entrants do not aim at understanding humans since deception and pretence are highly rewarded in this contest. Another criticism leveled against the Loebner Prize is that it confuses the Imitation Game with \textit{proof of human-like intelligence}. However, machines cannot reason like humans, as claimed by Searle in 1980 with his famous `Chinese Room experiment' \shortcite{searle1980minds,sep-chinese-room}.} whose results will be reported herein. 
For example, the virtual debater devised in \cite{slonim2021autonomous} exhibits a higher discussion quality than the compared artificial competitors, although it still fails to achieve a human-like level. Furthermore, \cite{balatsoukas2020wild} reported on the findings ensuing from the pilot study designed to assess a former version of the CONSULT system and the comprised chatbot. The outcome was a criticism concerning a lack of a more natural conversation flow when interacting with the bot. User studies have also been conducted to test the human-robot interaction presented within the ArgHRI system of \cite{sklar2015argumentation,azhar-sklar-ijrr:2017}. The results showed how argumentation-based dialogues contribute to enhancing trust towards the robots. Nonetheless, analysis of the dialogues themselves~\cite{sklar2018explanation} highlighted how the possibility of interrogating the bot to obtain explanations did not lead to a significant increase in performance from the human-robot team, nor a boost in user satisfaction. 

On the other hand, the SPA conversational agent introduced in \cite{rosenfeld2016strategical} outplayed the baseline chatbot (which harnessed a different, heuristic, strategy) when tested in its persuasion task, thus proving capable of delivering human-like level conversations. Similarly outperforming the baseline agent is the bot presented in \cite{chalaguine2019impact}. Indeed, the paper includes an experiment that shows how such a chatbot, by positing arguments that address the users'
concerns, is more likely to positively change the users' attitude in comparison with another agent that does not employ such a strategy. An analogous interest in users' concerns is encompassed in the study implemented in \cite{chalaguine2020persuasive}. The results (conjointly supported by the experiments in \cite{hadoux2019comfort} and confirmed by \cite{hadoux2021strategic}) conclude that a strategic chatbot accounting for concerns is more likely to provide relevant and cogent arguments. Moreover, it is also worth mentioning the evaluation outcome of the other two persuasive agents presented in \cite{andrews2008argumentative,chalaguine2021addressing}. The former bot provides fluent conversations with its interlocutors performing generally better than a purely task-oriented system. The latter, instead, shows how an interactive chatbot yields more compelling information than a static webpage.

Lastly, the ArgueBot conversational agent underwent both quantitative and qualitative assessments \shortcite{wambsganss2021arguebot}. The data collected from detailed feedback and Likert scale post-experiment forms yielded positive results. In particular, the participants perceived the chatbot as helpful, useful and easy to interact with.

\section{Discussion}\label{sec:discussion}

\begin{figure}[ht!]
\begin{center}
\includegraphics[width=0.9\textwidth]{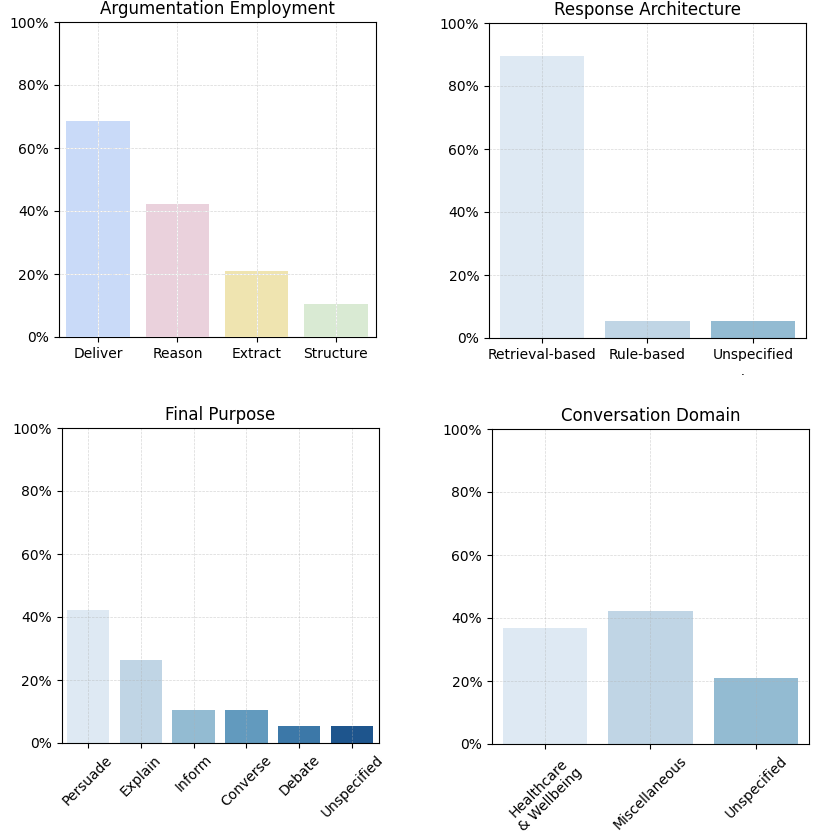}
\caption{\footnotesize{Percentage of sampled systems characterised by the argumentation employment type (top left), response architecture (top right), final purpose (bottom left) and conversation domain (bottom right) as described from the data of Table ~\ref{argumentation-based chatbots specifics}.}}
\label{fig:table2-stats}
\end{center}
\end{figure}

Table~\ref{argumentation-based chatbots specifics}  
depicts an overview of our findings, with a quantitative summary of the sampled chatbots' features shown in Figure~\ref{fig:table2-stats}. As a first remark, it is surprising that only a few argumentation-based chatbots appear in the literature. Indeed, the formal characterisation of real-world dialectical interactions provided by computational argumentation seems to be well-suited for agents whose role concerns conversing with users. This, however, may follow from the fact that the computational argumentation research field is still in an early stage of dissemination (especially outside of Europe), rather than deriving from the unsuitability of the argumentation formalism. Another possible explanation may be due to the fact that there has been an explosive interest in model-free methods in computer science in the last decades \cite{sep-artificial-intelligence}, ignoring model-based methods (like computational argumentation), which are only now gaining favour again, for example, as a way of `interpreting' the model-free output. Nevertheless, a number of considerations can be drawn from the outcome of our analysis.
\emph{Persuade} and \emph{Explain} prove to be the most common goals of the examined chatbots. The latter stems from the recent interest in explainable AI and its link with computational models of arguments \shortcite{vassiliades2021argumentation,mcburney2021argument,vcyras2021argumentative}. Persuasion dialogues, instead, have been studied in papers such as \shortcite{hunter2015modelling,murphy2016heuristic}, whose findings show how the use of argumentation-based formalisms may provide compelling strategies to induce belief change. 
One reason for such a number of persuasion-focused chatbots could indeed be related to the effectiveness of argumentation in delivering replies in such an area, as also advocated by the results of several user studies. To corroborate this, it can be noticed how persuasive conversational agents employ computational argumentation in such a way that falls under the (dialogical) \emph{Deliver} category (which, as expected, turns out to be the most common class listed in Table \ref{argumentation-based chatbots specifics}). Observe also that the main features of such bots include the account of beliefs and concerns when positing cogent (argumentative) replies. Continuing our analysis of different typologies of argumentation employment, it is worth emphasizing that \emph{Structure} always appears together with \emph{Reason} (though not vice versa), meaning that they are closely intertwined. That is because, in the considered papers, the instantiated scheme templates that structure the arguments work as input for the evaluating algorithm operated by the reasoning engine.

In general, it is less common that an argumentation-based chatbot employs argumentation solely for its reasoning engine. Indeed, after the semantics of the underlying AF have been computed, the bot usually requires a dialogue protocol that handles the replies delivery. Speaking of the underlying argumentation framework, we realized that,
when embedding a knowledge base into an AF, the Bipolar framework (and its variants QBAF and WBAF) turns out to be the most common option. This choice is related to the additional information provided by BAFs which encompass \emph{support} relations rather than just \emph{attacks}, allowing for an intuitive formalisation of both endorsements and conflicts between pieces of data. 

Within our survey, we identified several conversation domains contemplated by the bots, ranging from \emph{Healthcare} to \emph{Nuclear energy}, with the former representing the prevailing domain (and also subsuming others). Notice that `unspecified domain' could mean either that no conversational topic has been specified or that a sketched list of multiple topics has been presented. Interestingly, there is no argumentation-based chatbot eligible to be considered as open domain, although we might regard as `semi-open domain' the agent discussed in~\cite{slonim2021autonomous}. Indeed, despite the absence of topic limitations in its debate delivery (due to a huge corpus upon which arguments are retrieved), the bot is not capable of handling small talk or other analogous trivial interactions. This also affects its discussions, each of which is modelled as a challenge towards opposite stances. Another peculiarity of the agent engineered in~\cite{slonim2021autonomous} is that it allows for unconstrained speech in user input, whereas most chatbots only allow for free-text input (and the bots envisaged in~\cite{bex2016ai} and \shortcite{guo2022using} combines both free and limited textual prompts). Nonetheless, the proficiency in managing and processing unrestricted natural language sentences shows how argumentation-based chatbots can aptly mimic real-world-like discussions.

Finally, observe that almost every examined bot is equipped with a \emph{retrieval-based} response model with the only exception envisaged in \cite{cocarascu2019extracting}. Indeed, the hybrid conversational agent proposed therein handles its dialogues mostly via a few tailored textual templates, hence harnessing its \emph{rule-based} component. However, it may also resort to its retrieval-based model when the user questions the provided explanations. In general, it is also worth noticing that, unlike standard conversational agents, the surveyed literature revealed no \emph{generative-type} argumentation-based chatbots\footnote{Recall, however, that we have no explicit information regarding the response architecture of ArgueBot \shortcite{wambsganss2021arguebot}.}. Per se, this is not a major drawback, since generative response architecture may suffer from various issues such as lack of transparency about the origins of the produced replies, biased output, or creation of nonsensical responses. Nevertheless, this outlines a current limitation of argumentation-based bots, mostly due to an absence of studies on the matter. A possible solution to such a shortcoming may be provided, once again, by resorting to a hybrid approach that leverages state-of-the-art Transformer technologies. For example, embedding argumentation methodologies into current LLMs-based conversational agents would produce generative argumentation-based chatbots while also proving useful in mitigating those models' downsides. 

\subsection{Benefits of Leveraging Computational Argumentation Approaches in Generative-LLMs Chatbots Design}
In the literature, the class of generative-LLMs chatbots (e.g. ChatGPT, Llama 2-Chat, Bard, Claude) is considered to be the present cutting-edge category of conversational agents. Having already listed the shortcomings that affect those models, we have not yet discussed potential solutions on how to address such limitations. We argue that computational argumentation may prove to be an effective means capable of successfully handling and amending most of these weaknesses, especially (but not limited to) when they originate from the black-box nature of LLMs. Indeed, the thriving research field of \emph{eXplainable AI (XAI)}, which studies ways to improve the interpretability of AI-driven systems, proposes also argumentative strategies as adequate forms of explanations to address the lack of models' transparency \shortcite{vcyras2021argumentative,vassiliades2021argumentation}. These intuitions are backed by studies such as \cite{mcburney2021argument,castagna2022towards}, where it is suggested that AI systems should adopt an argumentation-based approach to explanations consisting of dialogue protocols characterising the interactions between an explainer and an explainee. Embedded into LLMs, such a dialectical interplay would provide an informative post hoc method to deliver deliberated explanations to end-users while also ensuring detailed replies to follow-on queries.

On this matter, it is worth noticing that Microsoft conducted an analysis of the capability of GPT-4 (one of the latest released GPT models \shortcite{openai2023gpt4}) to provide clarifications regarding its output \shortcite{bubeck2023sparks}. Although it outperforms the ChatGPT version based on GPT-3.5, even GPT-4 has its drawbacks when dealing with the \emph{process consistency} of its explanations: it provides a plausible account of the rationale behind the generation of its output, but it often fails in representing a more general justification able to predict the outcome of the model given similar inputs. An argumentative dialogue (such as EQR \cite{mcburney2021argument,castagna2022towards}) designed for explanation purposes would solve the process-consistency issues by providing conversations where more information can be retrieved and thus eschewing the limited explanation length and language constraints deemed to be the leading causes of the problem \shortcite{bubeck2023sparks}. 
\\\\Drawing from the usability of the aforementioned dialogue-based XAI, let us now delve into the possible ways in which computational argumentation may provide solutions (summarized in Table 3) to the current shortcomings of LLMs:
\begin{itemize} 
    \item[] \textbf{Emergent abilities.} The puzzling appearance of such an unpredictable phenomenon consists of the sudden occurrence of specific competencies in large-scale models that do not manifest in smaller ones. Thus, it is not possible to anticipate the `emergence' of these abilities (e.g. improved arithmetic, multi-task understanding, enhanced multi-lingual operations) by simply analysing smaller-scale models \shortcite{wei2022emergent}. Among these capabilities, we can also identify Theory of Mind (ToM), i.e. the aptitude to impute mental state to others. Considered to be uniquely human, ToM may have spontaneously occurred in LLMs as a byproduct of their training \shortcite{kosinski2023theory}. All of the aforementioned aspects contribute to the general mystery surrounding Transformer-based technology, which leads to mistrust among the general public. Argumentative XAI could indirectly help as a post hoc solution: although it cannot identify the reasons why emergent abilities originate, it could nonetheless provide explanations that would clarify their functioning.
    \item[] \textbf{Hallucination.} Defined as \emph{`the generated content that is nonsensical or unfaithful to the provided source content'} \shortcite{ziwei2023hallucination} the phenomenon of hallucination in natural language generation can be divided into \emph{intrinsic} and \emph{extrinsic}. The former refers to generated output that contradicts the source upon which the model was trained. The second, instead, represents an output that cannot be verified. The employment of an argumentation reasoning engine can reduce the intrinsic hallucination kind by stipulating that only grounded arguments (hence, members of conflict-free sceptical extensions) will be output by the chatbot. On the other hand, extrinsic hallucinations can be probed by argumentative XAI methods, thus ensuring, in the worst-case scenario, the retrieval of additional information over the produced content. 
    \item[] \textbf{Reasoning.} Different scholars argue that, although LLMs provide a good representation of language generation, they lack reasoning skills and logical thinking \shortcite{mahowald2023dissociating,bang2023multitask,frieder2023mathematical,thorp2023chatgptfun}. In an attempt to provide effective solutions, Chain and Tree of Thoughts (respectively, CoT and ToT) have been introduced to address such weaknesses. CoT consists of a prompting strategy that details a series of intermediate reasoning steps in order to achieve better performance in arithmetic, symbolic and commonsense inferences \shortcite{wei2022chain}. The limitations of this approach mostly concern the absence of a procedure to plan or analyse multiple reasoning paths before generating the output and this is exactly the enhancement yielded by ToT. Indeed, Tree of Thoughts frames each problem as a search over a tree, where each node is a partial solution \shortcite{yao2023tree}. Against these two options, we argue that endowing generative-LLM-based chatbots with a reasoning engine driven by computational argumentation may provide a more intuitive and cheaper alternative (e.g. it does not require expensive resources to be implemented, unlike ToT). Argumentative reasoning is particularly suited for models that parse, work and generate natural language. Recall that AFs are graphs whose edges represent paths determining the status of each node. Then, semantically computing an argumentation framework allows planning the most appropriate sequence of `thoughts' (arguments) to achieve the desired result. Such sequences account for divergent information, thus also mimicking and (potentially) outperforming the recent CCoT (Contrastive Chain of Thought) prompting technique, which mostly handles only one contrastive sample at a time \shortcite{chia2023contrastive}.  
    \item[] \textbf{Biased and Toxic Output.} Models have a tendency to reflect their training data, thus reproducing biased or toxic content that can harm the interacting user \shortcite{brown2020language}. This translates into the critical necessity of aligning LLMs towards human moral values, and even in this case, computational argumentation may prove useful to mitigate the problem. Indeed, a recent study investigates the use of computational argumentation as a tool for detecting unwanted bias in tabular data-driven binary classification decision-making systems \cite{waller2023ohaai}. The proposed method is model-agnostic and does not require access to labelled data or the specification of protected
characteristics. Notice also that the steadfast progress in the field of argument mining could ensure the provision of algorithms capable of precisely detecting biased and toxic arguments in the underlying dataset and filtering them out. This would allow for the reduction of harmful data upon which generative models will be trained. Another potential solution envisages leveraging argument schemes and their taxonomies. Specifically, the instantiation of AS from AI systems enables a semantically richer approach capable of enhancing and leading LLMs-generated text into more realistic and ethically constructive debates \cite{elfia2023ohaai}. 
\end{itemize}

\begin{table}[ht!]
\centering
\begin{tabular}{c|ccc}
\hline
    \textbf{Generative LLMs Chatbot} &  \multicolumn{3}{c}{\textbf{Potential Solutions}} \\
    \cline{2-4}
    \textbf{Shortcomings} & \multicolumn{1}{c|}{\emph{\textbf{Arg XAI}}} & \multicolumn{1}{c|}{\emph{\textbf{Arg Engine}}} & \emph{\textbf{AM} $\&$ \textbf{AS}}\\
    \hline
    \emph{Emergent Abilities} & \multicolumn{1}{c|}{\ding{51}} & \multicolumn{1}{c|}{}\\
    \hline
    \emph{Hallucination} &  \multicolumn{1}{c|}{\ding{51}} & \multicolumn{1}{c|}{\ding{51}}\\
    \hline
    \emph{Reasoning} & \multicolumn{1}{c|}{} & \multicolumn{1}{c|}{\ding{51}}\\
    \hline
    \emph{Biased and Toxic Output} & \multicolumn{1}{c|}{} & \multicolumn{1}{c|}{} & \ding{51}\\
    \hline
\end{tabular}
\label{table:arg solving LLM issues}
\caption{\footnotesize{Computational argumentation means for addressing LLMs chatbots' downsides. Arg XAI (Argumentative XAI) refers to explanation procedures based on computational argumentation strategies and tools. Arg Engine (Argumentative Engine) concerns the reasoning capabilities of engines driven by computational argumentation (Section \ref{sec:Arg Reasoning Engine}). Finally, AM indicates the Argument Mining operations of Section \ref{sec:AM}, whereas AS denotes the Argument Schemes structure of Section \ref{sec:AS}.}}
\end{table}

\section{Conclusion} \label{sec:conc}
Conversational agents and computational argumentation are intrinsically connected by their shared focus on dialectical interactions. Combining both subjects, in this paper, we have sifted through the literature to review and analyse the existing argumentation-based chatbots. Around $70\%$ of the bots we examined (recalling our constrained selection, as explained in Section~\ref{sec:meth}) employ computational models of arguments as a way of delivering their replies to interacting users, following specific dialogue protocols. This implies that argumentative formalism proves to be particularly effective when handling exchanges of information in natural language, especially if a persuasion goal is involved.
In addition, reasoning engines prove to be quite a common feature too. Harnessing argumentation extensions, those engines provide the rationale for selecting the most appropriate response to output, depending on the chosen semantics.   
Finally, unlike standard bots (i.e. non-argumentative ones), we discovered that there is no generative argumentation-based chatbot, nor an open-domain one, although there might be some ways of implementing such agents by embedding argumentation methodologies within LLM-driven conversational agents.  
Entangled with computational argumentation, chatbot design and their respective forthcoming progress, the research field of argumentation-based chatbots appears to have promising options to pursue in the coming years, including an interesting role to play in the recent Transformer-based turn of AI studies.

\bibliography{chatbotsurvey.bib}
\bibliographystyle{theapa}
\end{document}